\documentclass[journal]{IEEEtran}
\pdfoutput=1
\usepackage{cite}
\usepackage[pdftex]{graphicx}
\usepackage{amsmath}
\usepackage{amssymb}  
\usepackage{algorithmicx}
\usepackage{mathtools}
\usepackage{url}
\usepackage[table]{xcolor}
\usepackage{bbm}
\usepackage{threeparttable}
\usepackage{hyperref}
\usepackage{caption}
\usepackage{subcaption}
\usepackage{makecell}
\usepackage{pifont}

\DeclareMathOperator*{\argmax}{arg\,max}

\newcommand*{\prob}{\mathsf{P}}

\begin{document}
%
\title{When Vehicles \textit{See} Pedestrians with Phones:\\
A Multi-Cue Framework for Recognizing Phone-based Activities of Pedestrians}

\author{Akshay~Rangesh,~\IEEEmembership{Member,~IEEE,}
        and Mohan~M.~Trivedi,~\IEEEmembership{Fellow,~IEEE}
        \thanks{The authors are with the \href{http://cvrr.ucsd.edu/}{Laboratory for Intelligent and Safe Automobiles}, University of California, San Diego, CA 92092, USA.
        
email - {\tt\small arangesh, mtrivedi@ucsd.edu}}
}

\maketitle

\begin{abstract}

The intelligent vehicle community has devoted considerable efforts to model driver behavior, and in particular to detect and overcome driver distraction in an effort to reduce accidents caused by driver negligence. However, as the domain increasingly shifts towards autonomous and semi-autonomous solutions, the driver is no longer integral to the decision making process, indicating a need to refocus efforts elsewhere. To this end, we propose to study pedestrian distraction instead. In particular, we focus on detecting pedestrians who are engaged in secondary activities involving their cellphones and similar hand-held multimedia devices from a purely vision-based standpoint. To achieve this objective, we propose a pipeline incorporating articulated human pose estimation, followed by a soft object label transfer from an ensemble of exemplar SVMs trained on the nearest neighbors in pose feature space. We additionally incorporate head gaze features and prior pose information to carry out cellphone related pedestrian activity recognition. Finally, we offer a method to reliably track the articulated pose of a pedestrian through a sequence of images using a particle filter with a Gaussian Process Dynamical Model (GPDM), which can then be used to estimate sequentially varying activity scores at a very low computational cost. The entire framework is fast (especially for sequential data) and accurate, and easily extensible to include other secondary activities and sources of distraction.


\end{abstract}

\section{Introduction}

\IEEEPARstart{W}{ith} the explosion of hand-held device usage globally, smart phones have made their way into most hands. This trend is expected to continue as devices get cheaper and find more utility in our day to day lives. As of 2011, there were more phones than people in the USA, and internationally, the number of mobile phone subscriptions is an estimated 5.9 billion. Though such devices are extremely useful and even indispensable for many, it is this very dependence that is a major cause of pedestrian distraction, and possible injury. From here on-wards, we shall make use of the term \textit{cellphone} as a placeholder for any hand-held multimedia device that a pedestrian may interact with.

Distracted walking, like distracted driving, is likely to increase in parallel with the penetration of electronic devices into the consumer market. Although driver distraction has received abundant attention since the turn of the century, distraction among pedestrians is a relatively nascent area of research. This is surprising given that pedestrians are in fact prone to acting less cautiously when distracted. Furthermore, a recent report by the Governors Highway Safety Association (GHSA) reveals a disturbing trend - between the mid-1970s and early 2000s, pedestrian deaths steadily declined, eventually dipping to around 11 percent of all motor vehicle fatalities. But since 2009, pedestrian fatalities have actually increased by 15 percent, climbing to 4,735 in 2013. Meanwhile, the percentage of pedestrians killed while using cell phones has risen, from less than 1 percent in 2004 to more than 3.5 percent in 2010, according to\cite{nasar2013pedestrian}. Also, the study shows that the number of pedestrians injured while on their cellphones has more than doubled since 2005.

\begin{figure}[t]
\begin{center}
\includegraphics[width=0.95\linewidth]{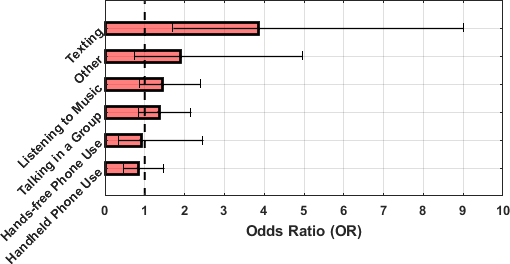}
\end{center}
\caption{Odds of failing to display optimal crossing behavior for different activities\cite{thompson2013impact}, along with their 95\% confidence intervals.}
\label{fig:OR}
\end{figure}

The severity of this phenomenon is further reflected by the number of studies conducted over the last few years, each of which arrive at similar conclusions. In a recent study conducted by Thompson et al.\cite{thompson2013impact}, they conclude that nearly one-third (29.8\%) of all pedestrians performed a distracting activity while crossing, with text messaging associated with the highest risk among different technological and social factors (Figure \ref{fig:OR}). Meanwhile, Nasar et al.\cite{nasar2013pedestrian} found that mobile-phone related injuries among pedestrians increased relative to total pedestrian injuries, and paralleled the increase in injuries for drivers, and in 2010 exceeded those for drivers. The study by Byington et al.\cite{byington2013effects} confirms this by a virtual street based simulation, stating that - while distracted, participants waited longer to cross the street, missed more safe opportunities to cross, took longer to initiate crossing when a safe gap was available, looked left and right less often, spent more time looking away from the road, and were more likely to be hit or almost hit by an oncoming vehicle. Moreover, it is noted that the demographic of individuals between ages 18 and 29 is more susceptible to exhibit such behavior. For a detailed report on the global nature of the pedestrian safety problem and the inadequacy of current systems in ensuring it, we refer the reader to\cite{gandhi2007pedestrian}.

It is also interesting to note that as the emphasis of automobile manufacturers gradually shifts towards more automated vehicles, so must the emphasis placed on preventing pedestrian distraction related injuries. In such scenarios, the intelligent vehicle must be able to gauge the risk associated with each pedestrian, and demonstrate more caution in avoiding those with larger risks.

In this study, we focus only on distraction due to technological factors, particularly the use of cellphones for different tasks, and ignore social impacts such as talking or walking in a group. To summarize, we propose to classify each of many pedestrians in an image, into one of 3 activity classes- \textit{none}, \textit{texting} and \textit{handheld phone call}. We additionally extend this approach to work on sequences of images, where knowledge about temporal dynamics can enable faster and more efficient operation.

The rest of the paper is organized as follows - Section \ref{related_work} briefly outlines some related work in the field. Section \ref{dataset_description} describes the data we are working with, and the semantic annotations that are available for use. Section \ref{single_frame} details the proposed methodology to estimate confidence scores of cellphone based activities for a pedestrian using a single image, and section \ref{sequential} extends this model to predict a score at every instance for a sequence of pedestrian images. Section \ref{experiments} lists the experiments carried out, and tabulates each of their results. Finally, section \ref{conclusions} concludes this work.

\section{Related Work}\label{related_work}
There is an abundance of work related to human activity recognition and classification from the last decade. However, these studies pertain to generic human activities and are not of much use in studying pedestrian distraction. Even though there have been quite a few studies that deal with driver distraction and activity modeling \cite{roth2016driver, hoang2016multiple, ohn2014head, tawari2014looking}, these models are not directly applicable to pedestrians because the forms of distraction and the activities of interest are considerably different. Nonetheless, there have been ample efforts devoted to studying pedestrians in the context of path prediction, intent analysis and action/activity recognition.
We briefly go over these tasks, highlighting how they differ from the goal of this work. This study may appear similar to our previous work \cite{rangesh2016pedestrians} in terms of the end goal, however, the proposed methodology is entirely different. We also use a significantly larger dataset and provide a more exhaustive evaluation in comparison to \cite{rangesh2016pedestrians}. For a more detailed list of studies conducted on humans around vehicles, we refer the reader to \cite{ohn2016looking}.

\textbf{Path prediction and gait analysis:} There have been numerous studies on predicting the trajectories of pedestrians to prevent collisions and improve surround vehicle safety. These methods generally ignore high-level semantics (such as pedestrian intent) and predict the paths based on low level cues alone \cite{trivedi2015trajectory, gandhi2008image, goldhammer2013early, prioletti2013part, bera2016glmp}.

\textbf{Intent analysis:} The aim of such studies is to make an estimate of the pedestrians' intention in the near future, so as to take appropriate measures to reduce risk of collision. These studies are commonly carried out in conjunction with path prediction, in a manner that benefits both tasks. Recent examples in this domain are \cite{kohler2012early, madrigal2014intention, schulz2015pedestrian, bandyopadhyay2013intention, keller2014will, kooij2014context}.

\textbf{Action/Activity recognition:} The terms action and activity have been used quite loosely in the context of pedestrians. In most cases, these terms allude to the different stages in the trajectory of a pedestrian \cite{kataoka2015fine, quintero2014pedestrian}, e.g. walking, waiting, crossing etc. This notion of activity has also been extended to groups of people, where portions of a crowd are assigned a common activity based on context and collective behavior \cite{choi2009they}. In this study, we use the term activity to refer to the \textit{secondary} activity of a pedestrian being performed in addition to walking/crossing.

Although, the tasks listed above are focused on modeling pedestrians and their behavior, none of them consider pedestrian distraction due to secondary activities like cellphone usage (see Table~\ref{table:related} for reference). Moreover, this study could be complementary to existing studies on pedestrian intention and path prediction, and could result in a more holistic understanding of pedestrian behavior.

\begin{table}[t!]
\setlength\arrayrulewidth{0.8pt}
\rowcolors{2}{white}{gray!25}
\centering
\caption{Related work in image based pedestrian safety.}
\begin{tabular}{| c | c | c | c |}
 \hline
 \thead{Study} & \thead{Path \\Prediction} & \thead{Intent \\Analysis} & \thead{Activity \\Recognition\\ (Output Classes)} \\
 \hline \hline
 Moeslund et al.\cite{trivedi2015trajectory} & \ding{51} & \ding{51} & - \\ 
 Gandhi et al.\cite{gandhi2008image} & \ding{51} & - & - \\
 Goldhammer et al.\cite{goldhammer2013early} & \ding{51} & \ding{51} & - \\
 K\"{o}hler et al.\cite{kohler2012early} & - & \ding{51} & - \\
 Madrigal et al.\cite{madrigal2014intention} & \ding{51} & \ding{51} & - \\
 Schulz et al.\cite{schulz2015pedestrian} & - & \ding{51} & \makecell{cross, turn into\\ road, stop} \\ 
 Bandyopadhyay et al.\cite{bandyopadhyay2013intention} & - & \ding{51} & - \\ 
 Keller et al.\cite{keller2014will} & \ding{51} & - & walking, stopping \\ 
 Kooij et al.\cite{kooij2014context} & \ding{51} & \ding{51} & walking, stopping \\
 Kataoka et al.\cite{kataoka2015fine} & - & \ding{51} & \makecell{crossing, walking, \\standing, riding \\a bicycle} \\
 Quintero et al.\cite{quintero2014pedestrian} & \ding{51} & - & \makecell{walking, starting,\ \ \\standing, stopping\ \ } \\
 Choi et al.\cite{choi2009they} & - & - & \makecell{collective activities \\of pedestrians \\like crossing, \\waiting, queuing, \\walking, talking} \\
 Rangesh et al.\cite{rangesh2016pedestrians} & - & - & \makecell{using phone, none}\\
 \textbf{This work} & - & - & \makecell{texting, handheld\ \ \ \\phone call, none\ } \\
 \hline
\end{tabular}
\label{table:related}
\end{table}

\begin{figure*}[!t]
    \centering
    \begin{subfigure}[t]{0.3\textwidth}
        \centering
        \includegraphics[height=1.1in]{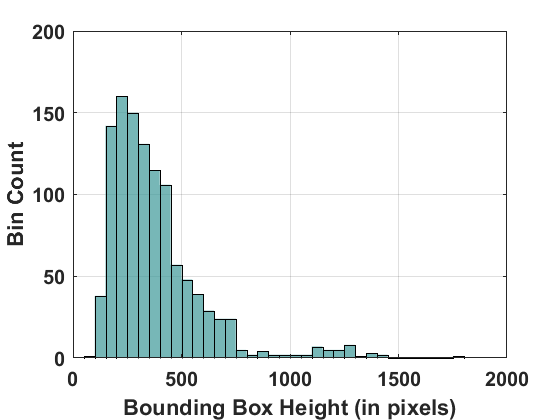}
        \caption{Histogram of pedestrian bounding box heights}
    \end{subfigure}%
    ~ 
    \begin{subfigure}[t]{0.3\textwidth}
        \centering
        \includegraphics[height=1.1in]{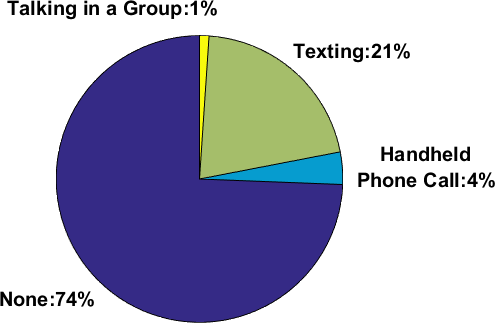}
        \caption{Pedestrian activities}
    \end{subfigure}%
    ~
    \begin{subfigure}[t]{0.3\textwidth}
        \centering
        \includegraphics[height=1.1in]{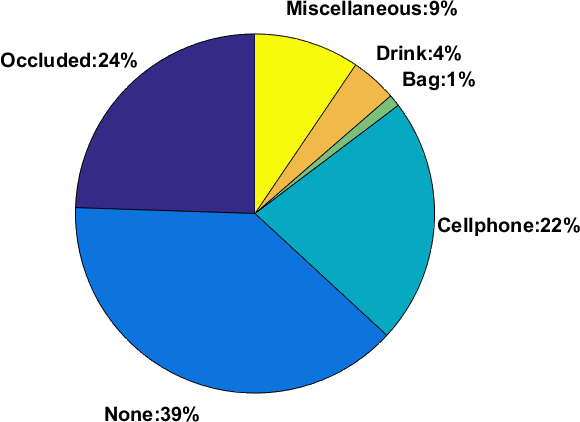}
        \caption{Objects in pedestrian hands}
    \end{subfigure}
    
    \begin{subfigure}[t]{\linewidth}
        \centering
        \includegraphics[height=1.05in]{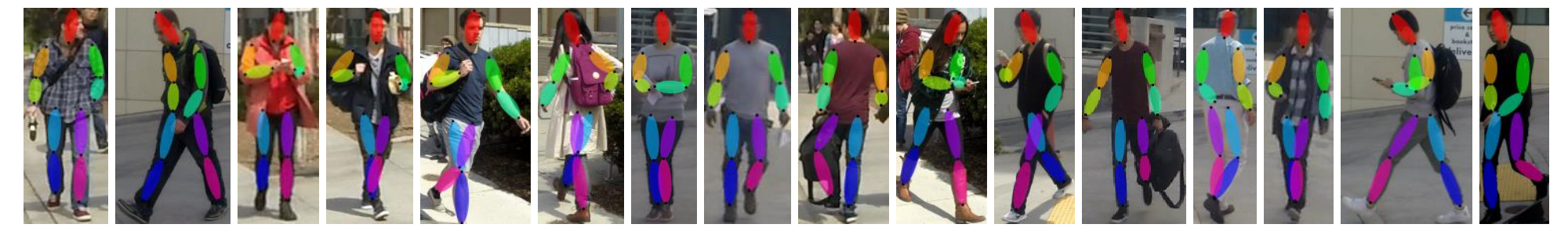}
        \caption{Pedestrian samples from the dataset. The joints obtained after articulated pose estimation have been overlaid for reference.}
    \end{subfigure}
    \caption{Details pertaining to the proposed dataset. The dataset is demonstrably diverse in viewpoints, pedestrian size, activity and object interactions.}
\label{fig:dataset}
\end{figure*}

\section{Dataset Description \& Semantic Annotations}\label{dataset_description}
Since pedestrian distraction due to cellphone usage is more common among a young demographic, we mounted 4 GoPro cameras, each facing a different direction, on an intelligent vehicle testbed parked at an intersection in the UC San Diego campus. By capturing different viewpoints on each camera, we ensure that pedestrians are not predisposed to appear in a particular location or facing a certain direction. Most of the data is captured on afternoons and evenings, on both sunny and overcast days to ensure diverse illumination conditions and reasonable foot traffic. Since the proposed methodology carries out fine-grained analysis of pedestrians, we avoid night time situations where it is hard to identify small objects and features even for humans. Furthermore, pedestrians are captured holding a variety of objects in addition to cellphones, such as bags, drinks, food and other miscellaneous items. To facilitate the finer analysis of each pedestrian, videos were captured at 2.7k resolution, resulting in pedestrians as large as 1000 pixels in height in a few cases. Figure~\ref{fig:dataset} visualizes certain key statistics of our dataset, and shows a few sample pedestrians chosen at random.

The dataset comprises of a total of 1586 cropped pedestrians, each with annotated activities and objects. These pedestrians are then divided into train and test sets using a 75-25 split, while making sure that the fraction of occurrences of each activity is retained in both sets.

Additionally, we annotate 7 sequences of pedestrians (3 for training, 4 for testing), each approximately 10 seconds in duration ($\approx300$ frames). In this case, the pedestrian is assigned an activity for each frame to account for temporal dynamics. In addition to this, the upper body joints (listed in section \ref{single_frame}) are annotated for each frame, to enable evaluation of the proposed articulated pose tracker.

\section{Single Frame Activity Classification}\label{single_frame}

\begin{figure}[t]
\begin{center}
\includegraphics[width=0.7\linewidth]{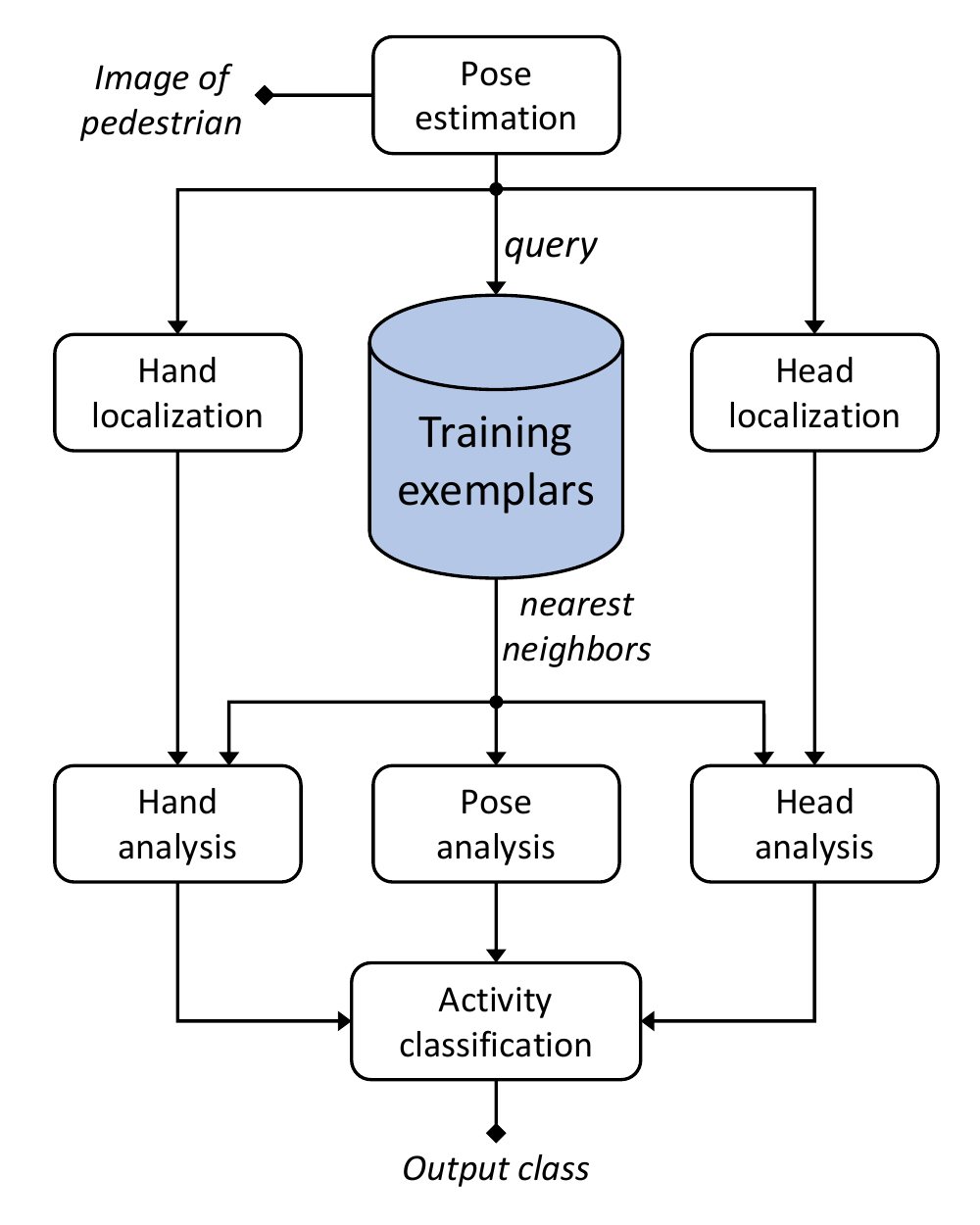}
\end{center}
\caption{Flow diagram of proposed methodology for single frame activity classification.}
\label{fig:pipeline}
\end{figure}

Figure~\ref{fig:pipeline} depicts the flow diagram of the proposed activity classification framework. The pipeline takes in an image patch corresponding to a pedestrian, and outputs the corresponding activity. We detail each processing block in the subsections that follow.

\subsection{Articulated Pose Estimation} \label{pose}
The articulated pose of a pedestrian can be an invaluable cue in estimating the activity he/she is involved in. Recent advances in pose estimation using deep convolutional neural networks (ConvNets) have led to state of the art results on challenging benchmarks. We make use of one such architecture, called the Convolutional Pose Machines\cite{wei2016convolutional} proposed by Wei et al. This is a multi-stage ConvNet, where each subsequent stage operates
both on image evidence as well as belief maps from preceding stages, gradually refining the pose estimate. This setup offers us great flexibility while choosing the number of stages, with the trade-off being speed versus accuracy. The network has been trained on the MPII dataset comprising of 25K images containing over 40K people, involved in 410 different activities, and outputs the locations of 16 joints corresponding to the articulated pose of a human body. We use this pre-trained network and fine-tune it on our own dataset. This gives us marginal improvements in performance compared to an out-of-the-box implementation (see Table~\ref{table:pose_single}). Additionally, we only make use of the upper body joints for any further processing, as these are the most informative in our application. The framework can easily accommodate the full body pose instead, if necessary. The final set of keypoint locations retained are - head, neck, left shoulder, left elbow, left wrist, right shoulder, right elbow and right wrist. See Figure~\ref{fig:dataset} for some visual results of the pose estimation module on the proposed dataset.

\begin{table}[t!]
\centering
\tabcolsep=0.09cm
\caption{PCK scores~\cite{yang2013articulated} of the pose estimation module \cite{wei2016convolutional} on the test set before and after fine-tuning on the train set. Values close to 1 indicate near-perfect keypoint localization.}
 \begin{tabular}{| c | c | c | c | c | c | c | c | c |} 
 \hline
 \thead{} & \thead{head} & \thead{neck} & \thead{right \\shoulder} & \thead{right \\elbow} & \thead{right \\wrist} & \thead{left \\shoulder} & \thead{left \\elbow} & \thead{left \\wrist} \\ [0.5ex] 
 \hline\hline
 \thead{Before\\fine-tuning} & 1.00 & 1.00 & 0.99 & 1.00 & 0.96 & 1.00 & 1.00 & 0.94\\ 
 \thead{After\\fine-tuning} & 1.00 & 1.00 & 0.99 & 1.00 & 0.97 & 1.00 & 1.00 & 0.96\\ 
 \hline
 \end{tabular}
 \label{table:pose_single}
\end{table}

Most human pose estimation algorithms require the rough location and scale of the human in the image plane. In this study, we assume that such information is available beforehand, and focus our attention on analyzing each pedestrian in finer detail. However, if desired, the location and scale of pedestrians may be obtained easily from any generic pedestrian detector. We would also like to point out that many recent studies like \cite{cao2016realtime, papandreou2017towards} demonstrate state-of-the-art multi-person pose estimation in real time, without prior information on pedestrian locations and scales. This makes our approach viable for time critical applications like pedestrian safety and path planning.

The pose estimation module is used in our pipeline for three specific purposes. First, it allows us to localize the head and hands of each pedestrian for further examination. Second, it is used to identify similar training exemplars in the pose space. Third, the pose alone may be used as a informative prior over all activities. In each of the following subsections, we make use of the articulated pose in a manner mentioned above.

\subsection{Hand Analysis using Exemplar SVMs} \label{esvm}

An important cue for predicting the activity of a pedestrian are the objects they interact with. To identify the objects held in the hands of a pedestrian, we look at local image patches around the location of each hand. To do so, we first regress to the approximate location of the hands of a pedestrian, assuming that it is \textit{collinear} with the joints corresponding to the elbow and wrist. Let $(x_e, y_e)$ and $(x_w, y_w)$ denote the image plane coordinates of the elbow and wrist respectively. Using the assumption above, the approximate location of the hand $(x_h, y_h)$ is obtained as follows:

\begin{equation}
x_h = x_e + \frac{x_w - x_e}{r},
y_h = y_e + \frac{y_w - y_e}{r},
\end{equation}
where $r$ is a a parameter that depends on the ratio of distances of the elbow from the wrist and hand respectively. In our experiments, $r = 5/6$ seemed to generate the best results.

\begin{figure}[!t]
  	\centering
	\begin{subfigure}[b]{0.2\textwidth}
		\centering
		\captionsetup{justification=centering}
		\includegraphics[width=\textwidth]{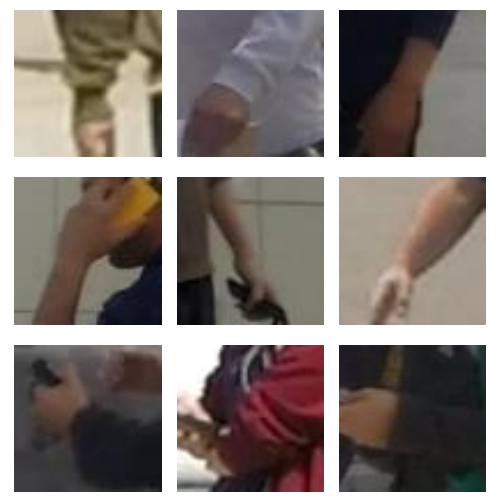}
		\caption{}
	\end{subfigure}%
~
	\begin{subfigure}[b]{0.2\textwidth}
		\centering
		\captionsetup{justification=centering}
		\includegraphics[width=\textwidth]{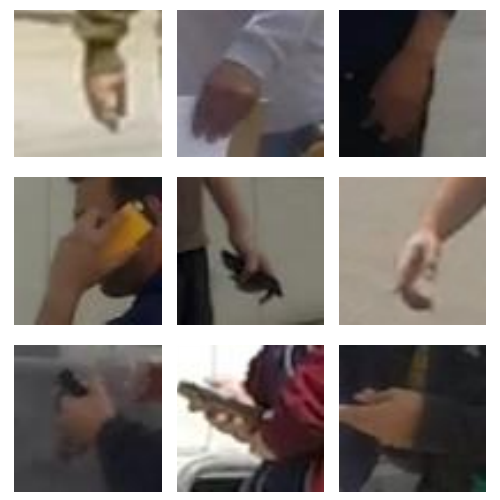}
 		\caption{}
	\end{subfigure}
\caption{Image patches obtained when the local window is centered around the (a)wrist versus the (b)hand.}
\label{fig:hands}
\end{figure}

Once we have the rough locations of both hands in the image plane, we crop out a local image window around these locations. The window size is chosen to be $\alpha \cdot h$ for a pedestrian parametrized by $(x, y, w, h)$. Here $\alpha$ is a hyper-parameter that ensures that the local window scales with the size of the pedestrian. In our experiments, $\alpha$ is set to $0.1$ to extract training patches. $\alpha$ is chosen to ensure that the hand almost always falls into the window, and also that the window is small enough to capture only the object of interest and nothing more. As demonstrated in Table~\ref{table:hand_loc}, $\alpha = 0.1$ offers the best results, beyond which increasing $\alpha$ does not improve hand localization by much. Examples of such local patches for windows centered around both the wrist and the hand can be found in Figure~\ref{fig:hands}. It is obvious that inferring the hand location, even if approximate, helps in centering the object of interest with respect to the window.

With a collection of such training patches centered around the hand, we proceed to build an object classifier. Our experiments demonstrated that traditional one-versus-all classifiers severely overfit the training data and failed to generalize well to new object instances. Moreover, training a separate classifier for each object class, as well as the intra-class variance (cellphones come in a variety of shapes and sizes) makes the classification task an especially hard one, considering the limited availability of training data.

We bypass all these limitations by training an ensemble of exemplar-SVMs (ESVMs) \cite{malisiewicz2011ensemble}. The method is based on training a separate linear SVM classifier for every exemplar in the training set. Each ESVM is thus defined by a single positive instance and millions of negatives, obtained by hard negative mining. In our case, an ESVM is trained to represent a rigid HOG template from an image patch around each hand of every pedestrian in the training set. At test time, the ESVM that results in the highest score is considered to provide the \textit{best match}, and the object label associated with the exemplar is transferred to the new test instance. Figure \ref{fig:matches} shows a few examples of matched hand-object instances.

\begin{table}[t!]
\centering
\tabcolsep=0.11cm
\caption{Fraction of pedestrian hands falling within predicted wrist and hand centered windows for different values of window scale factor $\alpha$. These evaluations were carried on a separate validation set.}
 \begin{tabular}{|c | c | c|} 
 \hline
 $\mathbb{\alpha}$ & \thead{Wrist-centered\\window} & \thead{Hand-centered\\window}\\
 \hline\hline
 0.05 & 0.0825 & 0.7250 \\ 
 0.07 & 0.3400 & 0.9275 \\
 0.10 & 0.9125 & 0.9900 \\
 0.12 & 0.9850 & 0.9950 \\
 0.15 & 0.9975 & 0.9975 \\
 0.20 & 0.9975 & 0.9975 \\
 \hline
 \end{tabular}
 \label{table:hand_loc}
\end{table}

\subsection{Gaze Analysis} \label{gaze}
The rough gaze direction of a pedestrian can be very effective in separating out instances where pedestrians are just holding a phone, versus when they are actually \textit{engaged} in its use.

\begin{figure*}[t]
\begin{center}
\includegraphics[width=\linewidth]{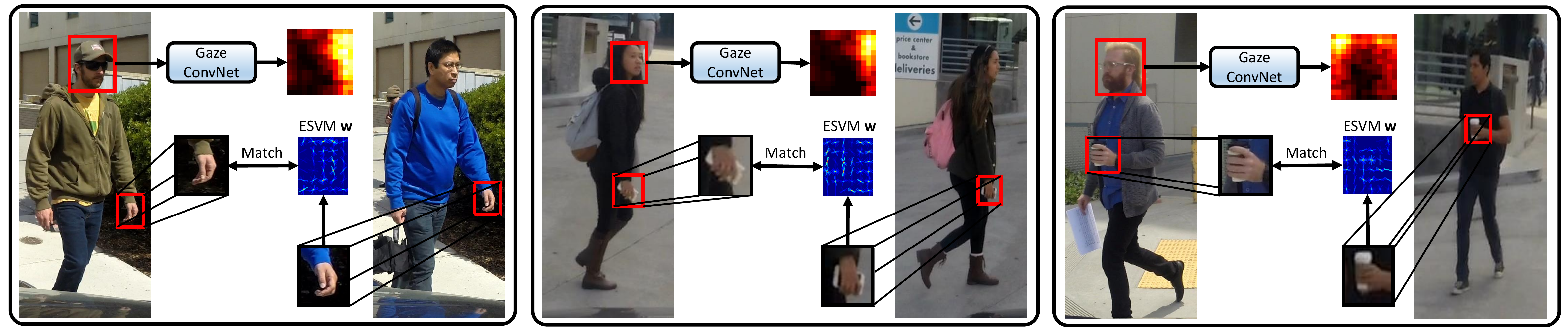}
\end{center}
\caption{Illustration of head and hand related cues described in sections \ref{esvm} and \ref{gaze}. In each of the three examples above, for a pedestrian in the test set (left), the gaze heatmap obtained from the gaze ConvNet is shown on top and the best hand (object) exemplar match with a pedestrian from the train set is shown below. The matched pedestrian (right) and exemplar weights are shown in addition to the matched hand patch. Best viewed in color.}
\label{fig:matches}
\end{figure*}

In this study, we use the \textit{gaze pathway} from the GazeFollow Convnet proposed in \cite{recasens2015they}. The gaze pathway takes in an image patch of the head along with its normalized location in the image plane (obtained from the articulated pose), and returns a $13 \times 13$ heat-map (Figure \ref{fig:matches}) that encodes the rough gaze direction of the pedestrian. This sub-network has five convolutional layers followed by three fully-connected layers, the final output of which is a single channel heat-map. Finally, we reshape this output to produce a 169-length feature vector that encodes the gaze.

\subsection{Querying Nearest Neighbor Pose Exemplars} \label{NN}
The main intuition behind our approach is that pedestrians with \textit{similar} body poses tend to interact with objects in a similar form, and are likely to be involved in analogous activities. To have such a notion of similarity, it is necessary to construct a suitable feature representation of the articulated pose, and to enforce a reasonable distance metric that ensures that similar poses are close by.

We make use of a combination of the normalized joint locations and the normalized joint angles as the feature descriptor. Consider a pedestrian bounding box parametrized as $(x, y, w, h)$. Here, $x$ and $y$ correspond to image coordinates of the top left corner of the bounding box, and $w$ and $h$ describe the dimensions of the box. For the pedestrian under consideration, the pose estimation network outputs a set of image locations $\{\mathbf{x}_i = (x_i, y_i)\}_{i=1,\cdots, 8}$ corresponding to each joint in the upper body. The set of normalized joint locations $\{\mathbf{\bar{x}}_i = (\bar{x}_i, \bar{y}_i)\}_{i=1,\cdots, 8}$ are then found as follows:

\begin{equation}
\bar{x}_i = \frac{x_i - x}{w},
\bar{y}_i = \frac{y_i - y}{h}, \forall i=1, \cdots, 8.
\end{equation}

Next, consider the set of joint triplets that are connected consecutively in the articulated pose tree. For each such triplet $(\mathbf{x}_i, \mathbf{x}_j, \mathbf{x}_k)$, let the angle subtended (in radians) at $\mathbf{x}_j$ by the line segment joining points $\mathbf{x}_i$ and $\mathbf{x}_k$ be denoted by $\angle \mathbf{x}_{(i, j, k)}$. We have 7 such joint angles in the upper body pose. The normalized joint angle at $\mathbf{x}_j$ is then obtained as follows:

\begin{equation}
\angle \mathbf{\bar{x}}_{(i, j, k)} = \frac{\angle \mathbf{x}_{(i, j, k)}}{\pi}.
\end{equation}

The final feature vector $\mathbf{x} \in \mathcal{R}^{23}$ is obtained as a simple concatenation of the set of normalized joint locations and angles. Our experiments indicated this to be much more stable in terms of closest neighbors in comparison to using either just the joint locations, or just the joint angles. With a set of pose features gathered from the pedestrians in the training set, we train a simple $K$-nearest neighbor classifier using a $K$-d tree structure for fast neighbor retrieval.

\subsection{Pedestrian Activity Classification}
Having set up the individual parts, we now focus on integrating the cues from the different modalities to predict a final class probability score. For this study, the possible output classes for activity classification are \textit{none}, \textit{texting} and \textit{handheld phone call}, which we encode as $y = 0, 1$ and $2$ respectively.

Consider a new pedestrian with pose features $\mathbf{x}_o$ calculated in the manner described above. The aim now is to predict a class label $y_o$, and estimate the probability associated with this prediction. Let $\mathcal{N}_K$ denote the set of $K$ nearest neighbor pose exemplars obtained from the trained classifier in \ref{NN}. We denote this set as follows - 

\begin{equation}
\mathcal{N}_K = \{(e_i^l, o_i^l, e_i^r, o_i^r, \mathbf{h}_i, y_i)\}_{i = 1, \cdots, K},
\end{equation}
where $e_i^l$ and $e_i^r$ denote the trained ESVMs on the left and right hands (from \ref{esvm}), $o_i^l$ and $o_i^r$ denote the object labels associated with the left and right hands, $\mathbf{h}_i$ represents the gaze features obtained as mentioned in \ref{gaze}, and $y_i$ denotes the activity label associated with the $i^{th}$ nearest neighbor exemplar.

Let $\mathcal{E}$ denote the image evidence available for the pedestrian whose activity is to be predicted. The desired predictive distribution may then be expressed as - 

\begin{equation}
\prob(y_o | \mathcal{E}; \mathcal{N}_K) \propto \prob(\mathcal{E} | y_o; \mathcal{N}_K)\ \prob(y_o; \mathcal{N}_K).
\end{equation}

Decomposing the image evidence into individual head and hand based evidences $\mathcal{E}_{head}$ and $\mathcal{E}_{hand}$, and making use of conditional independence yields - 

\begin{equation} \label{eq:eq6}
\prob(y_o | \mathcal{E}; \mathcal{N}_K) \propto \prob(\mathcal{E}_{gaze} | y_o; \mathcal{N}_K)\ \prob(\mathcal{E}_{hand} | y_o; \mathcal{N}_K)\ \prob(y_o; \mathcal{N}_K).
\end{equation}

Each term in the equation above is described below - 

\begin{equation} \label{eq:eq7}
\prob(\mathcal{E}_{gaze} | y_o; \mathcal{N}_K) \coloneqq \max\limits_{1 \leq i \leq K} \bigg(\frac{\mathbf{h}_o \cdot \mathbf{h}_i}{||\mathbf{h}_o||_2 \cdot ||\mathbf{h}_i||_2}\bigg) \cdot \mathbbm{1}_{\{y_o\}}(y_i),
\end{equation}
where $\mathbf{h}_o$ and $\mathbf{h}_i$ are the gaze descriptors, and $\mathbbm{1}_\mathcal{A}(\cdot)$ is the indicator function for set $\mathcal{A}$. This is simply the cosine similarity between the gaze features within the same class.

Next, let us denote the maximum match score obtained for ESVM $e_i^l$ on the left hand image patch as $p_i^l$, and that for $e_i^r$ on the right hand image patch as $p_i^r$. This probabilistic score is obtained by testing each ESVM on the corresponding image patch, and then re-scaling the match score using the parameters determined by carrying out the Platt calibration for each ESVM offline. Further, only matches with an overlap score greater than 0.4 with the test patch are retained as done in \cite{shrivastava2011data}. We can now define the hand evidence likelihood as follows - 

\small
\begin{equation} \label{eq:eq8}
\prob(\mathcal{E}_{hand} | y_o; \mathcal{N}_K)
 \coloneqq 
\begin{cases}
	\min\ \bigg(\max\limits_{i}\ p_i^l \cdot \mathbbm{1}_{\{0\}}(y_i), \\
	\max\limits_{i}\ p_i^r \cdot \mathbbm{1}_{\{0\}}(y_i)\bigg),\ \ \ \ \ \ \ \ \ \  \text{if $y_o = 0$}\\
	\\
	\max\limits_{i}\ \bigg(\big[p_i^l \cdot \mathbbm{1}_{\{1\}}(o_i^l) + \\ 
p_i^r \cdot \mathbbm{1}_{\{1\}}(o_i^r)\big] \cdot \mathbbm{1}_{\{y_o\}}(y_i)\bigg),\ \ \text{if $y_o = 1, 2$},
\end{cases}
\end{equation}
\normalsize
where $\mathbbm{1}_{\{1\}}(o_i^l)$ equals $1$ if the object associated with the left hand is a cellphone, else it equals 0. The same is true for $\mathbbm{1}_{\{1\}}(o_i^r)$ and the right hand.

Finally, the term $\prob(y_o; \mathcal{N}_K)$ acts as a prior over the activities, given just the articulated pose of a pedestrian. This is defined to be - 

\begin{equation} \label{eq:eq9}
\prob(y_o; \mathcal{N}_K) \coloneqq \frac{\sum\limits_{i = 1}^{K} \mathbbm{1}_{\{y_o\}}(y_i)}{K}.
\end{equation}

Using the equations \ref{eq:eq6}-\ref{eq:eq9}, the final predicted activity for the pedestrian is then chosen to be the MAP estimate - 

\begin{equation}
y_o^* = \argmax\limits_{y \in \{0, 1, 2\}}\ \prob(y | \mathcal{E}; \mathcal{N}_K).
\end{equation}

Since the probability terms on the right hand side of equation~\ref{eq:eq6} are not calibrated to provide compatible scores, we propose a second method based on late fusion of these scores. To do so, we create 9-length score vectors made up of the terms $\prob(\mathcal{E}_{gaze} | y_o; \mathcal{N}_K)$, $\prob(\mathcal{E}_{hand} | y_o; \mathcal{N}_K)$ and $\prob(y_o; \mathcal{N}_K)$ for $y_o \in \{0, 1, 2\}$. These vectors are created by performing a 5 fold cross validation split on the training set. Multi-class classification is carried out in a one versus all manner to predict the final activity of the pedestrian.

\section{Activity Classification for Sequential Data}\label{sequential}
In our proposed framework, the major bottleneck in terms of speed is the pose estimation network described in \ref{pose}. Even though it is possible to reliably estimate the pose for a few pedestrians in real time using a GPU (for a reasonable number of stages in the network), the network can no longer operate at a desired frequency when the number of pedestrians in the scene are considerably large. This issue can be alleviated by tracking the articulated pose of pedestrians for the duration between successive outputs from the pose estimation network. This also ensures that the pose estimated by the network makes reasonable transitions between successive instances, thereby reducing single frame errors.

\subsection{GPDM-based Particle Filter for Articulated Pose Tracking}
In this sub-section, we briefly describe the proposed particle filter based tracking framework with a Gaussian Process Dynamical Model (GPDM)\cite{wang2005gaussian}. Let $\mathbf{x}_t^i \in \mathcal{R}^{23}$ be the state of particle $i$ at time $t$, which represents the normalized pose features of a pedestrian as described in \ref{NN}. Let $\mathbf{l}_t^i \in \mathcal{R}^2$ denote the latent space projection of $\mathbf{x}_t^i$ using a Gaussian Process Latent Variable Model (GPLVM) as described below - 

\begin{equation} \label{eq:eq11}
\mathbf{x}_t = f(\mathbf{l}_t ; \mathbf{A}) + \mathbf{n}_{x, t}.
\end{equation}

In addition to this, a GPDM enforces an auto-regressive dynamical model in the latent space - 

\begin{equation} \label{eq:eq12}
\mathbf{l}_t = g(\mathbf{l}_{t-1} ; \mathbf{B}) + \mathbf{n}_{l, t}.
\end{equation}
Here, $\mathbf{n}_{x, t}$ and $\mathbf{n}_{y, t}$ are zero-mean, white Gaussian processes, $f$ and $g$ are nonlinear mappings parametrized by $\mathbf{A}$ and $\mathbf{B}$ respectively. Using small training sequences $(\mathbf{x}_0, \mathbf{x}_1, \cdots, \mathbf{x}_T)$, we can solve for both the corresponding latent space projections $(\mathbf{l}_0, \mathbf{l}_1, \cdots, \mathbf{l}_T)$, and the necessary hyperparameters in closed form\cite{wang2005gaussian}. Despite the use of small data sets, the GPDM learns an effective representation of the highly non-linear dynamics associated with articulated pose tracking.

At any instant $t$, the particle filter functions by propagating a set of particles in the latent space $\{\mathbf{l}_t^i\}_i$, by sampling (with noise) from the dynamical model in equation \ref{eq:eq12}. This results in an updated set of particles $\{\mathbf{l}_{t+1}^i\}_i$. To determine the likelihood of each particle, it is necessary to project the latent particles back into the observation space using the learned GPLVM mapping (equation \ref{eq:eq11}), where they may be evaluated against available measurements. This results in a corresponding set of particles $\{\mathbf{x}_{t+1}^i\}_i$ in the observation space. To evaluate the likelihood of each particle, we note that most pose estimation networks output a \textit{heatmap} for each joint location, which can be interpreted as a probabilistic score for its location in the image. With this in mind, let $h_j(\cdot, \cdot)$ denote the heatmap for the $j^{th}$ joint; the function takes in the $x$ and $y$ coordinates of any location on the image plane, and returns the score associated with the location, encoded in the heatmap. The likelihood of a particle $\mathbf{l}_{t+1}^i$ is then considered to be - 

\begin{equation}
\mathcal{L}(\mathbf{l}_{t+1}^i) = \mathcal{L}(\mathbf{x}_{t+1}^i) \coloneqq \prod\limits_{j = 1}^{8} h_j(x_j^i, y_j^i),
\end{equation}
where $(x_j^i, y_j^i)$ denotes the co-ordinates of the $j^{th}$ joint obtained from the pose features $\mathbf{x}_{t+i}^i$.

\begin{figure}[t!p]
    \centering
    \begin{subfigure}[b]{0.2\textwidth}
        \centering
        \includegraphics[height=1in]{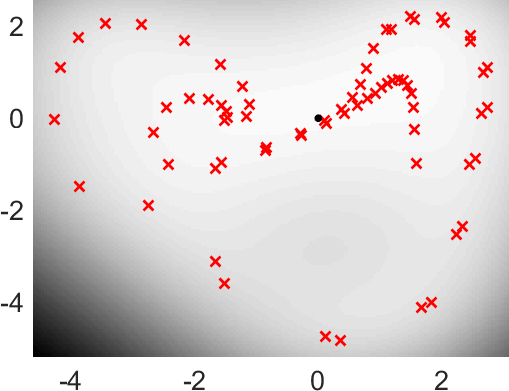}
        \caption{}
    \end{subfigure}%
    \begin{subfigure}[b]{0.2\textwidth}
        \centering
        \includegraphics[height=1in]{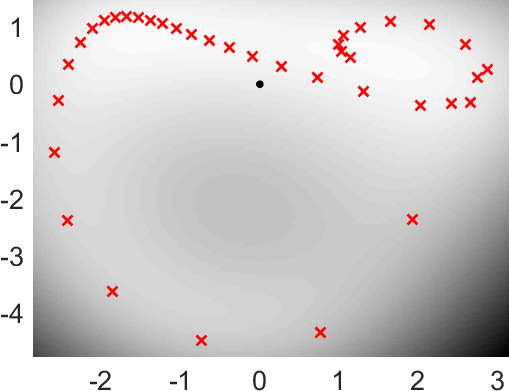}
        \caption{}
    \end{subfigure}

    \begin{subfigure}[b]{0.2\textwidth}
        \centering
        \includegraphics[height=1in]{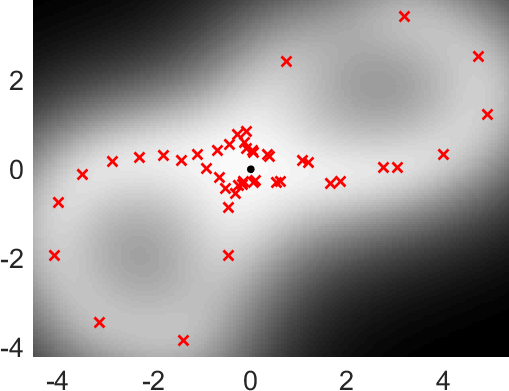}
        \caption{}
    \end{subfigure}%
    \begin{subfigure}[b]{0.2\textwidth}
        \centering
        \includegraphics[height=1in]{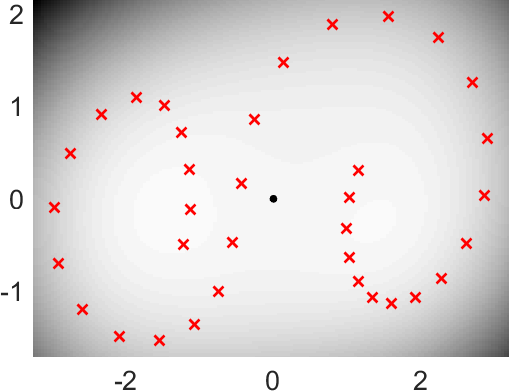}
        \caption{}
    \end{subfigure}
    \caption{Latent space projections of articulated pose trajectories for four different viewpoints.}
\label{fig:GPLVM}
\end{figure}

In practice, we train a set of GPDMs for different activities and viewpoints (e.g. walking towards the camera, walking away from the camera, walking sideways etc.). During test time, particles are initialized in latent space by locating the latent point (across all GPDMs) whose mapping in observation space yields the best match with the current measurement. Figure \ref{fig:GPLVM} shows the learnt latent space mappings for 4 different viewpoints.

\subsection{Avenues for Speedup}\label{speedup}
The tracking framework considerably reduces the burden on the pose estimation network in ensuring near real time operation. When a large number of pedestrians are present in the scene, we can simply run the networks on a subset of pedestrians, while the poses of the rest are updated based on state updates alone. When the network is finally run on a given pedestrian, the \textit{heatmaps} are used as measurements to update the state of each particle in the filter. Alternatively, one can make use of more recent real-time algorithms for multi-person pose estimation \cite{cao2016realtime, papandreou2017towards}, which enables faster operation due to reduced overhead from having separate detection and pose estimation modules.

Additionally, by only running the ESVMs associated to the K nearest neighbor exemplars, we bypass the computational drawbacks associated with ESVMs. We can further reduce the computational burden required to predict class labels for a pedestrian at every instant in case of sequential data. Once the hand evidence term $\prob(\mathcal{E}_{hand} | y_o; \mathcal{N}_K)$ in equation \ref{eq:eq8} is reliably estimated for all output classes, we need only update the head evidence and prior terms in equation \ref{eq:eq6} at every instant. This removes the need to run the ensemble of ESVMs at every frame.

\begin{figure*}[t!]
  	\centering
	\begin{subfigure}[b]{0.22\textwidth}
		\centering
		\captionsetup{justification=centering}
		\hspace{0.1in} \caption{\ \ $K = 25$}
		\includegraphics[width=\textwidth]{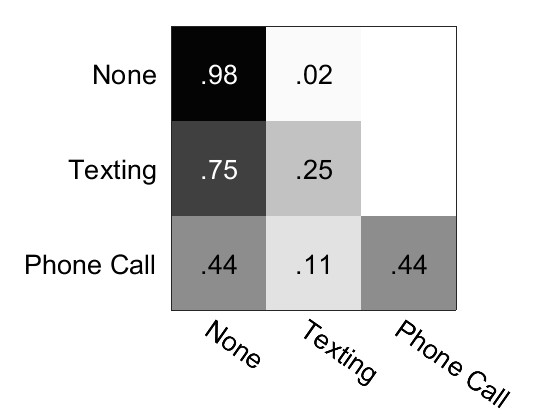}
	\end{subfigure}%
	~
	\begin{subfigure}[b]{0.22\textwidth}
		\centering
		\captionsetup{justification=centering}
		\hspace{0.1in} \caption{\ \ $K = 50$}
		\includegraphics[width=\textwidth]{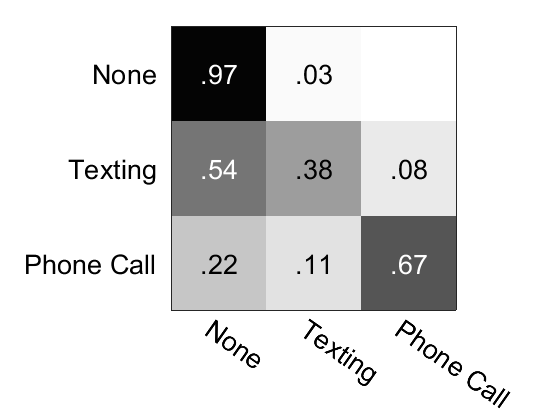}
	\end{subfigure}
	~
	\begin{subfigure}[b]{0.22\textwidth}
		\centering
		\captionsetup{justification=centering}
		\hspace{0.1in} \caption{\ \ $K = 100$}
		\includegraphics[width=\textwidth]{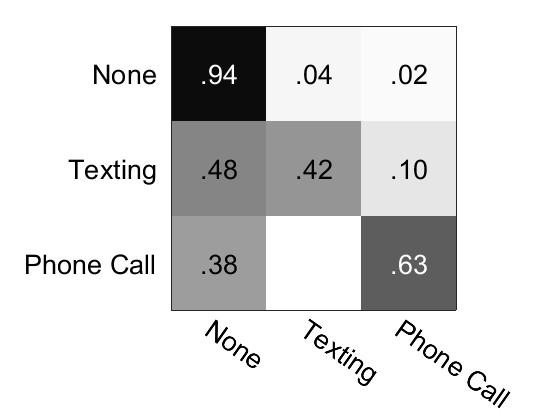}
	\end{subfigure}%
	~
	\begin{subfigure}[b]{0.22\textwidth}
		\centering
		\captionsetup{justification=centering}
		\hspace{0.1in} \caption{\ \ $K = 200$}
		\includegraphics[width=\textwidth]{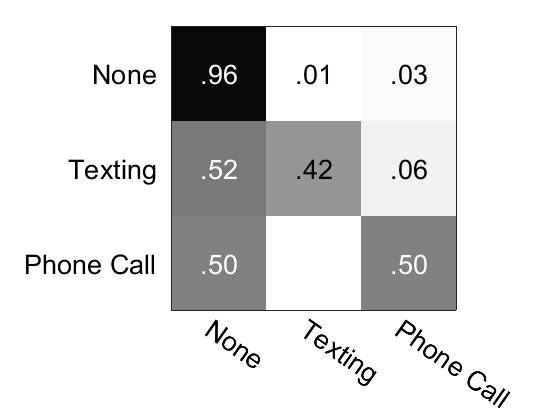}
	\end{subfigure}
	\vspace{0.1in}
	
	\begin{subfigure}[b]{0.22\textwidth}
		\centering
		\captionsetup{justification=centering}
		\hspace{0.1in} \caption{\ \ $K = 25$}
		\includegraphics[width=\textwidth]{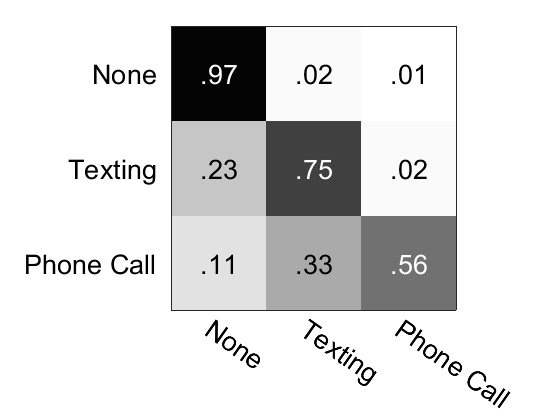}
	\end{subfigure}%
	~
	\begin{subfigure}[b]{0.22\textwidth}
		\centering
		\captionsetup{justification=centering}
		\hspace{0.1in} \caption{\ \ $K = 50$}
		\includegraphics[width=\textwidth]{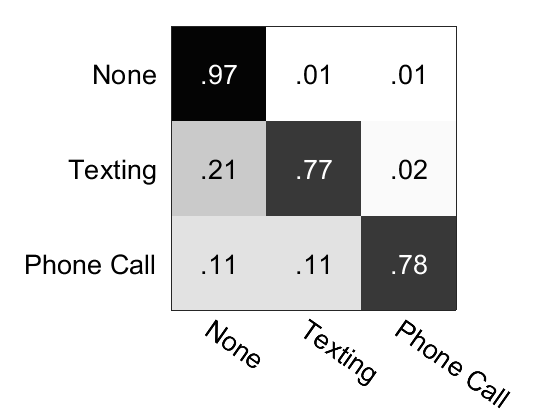}
	\end{subfigure}
	~
	\begin{subfigure}[b]{0.22\textwidth}
		\centering
		\captionsetup{justification=centering}
		\hspace{0.1in} \caption{\ \ $K = 100$}
		\includegraphics[width=\textwidth]{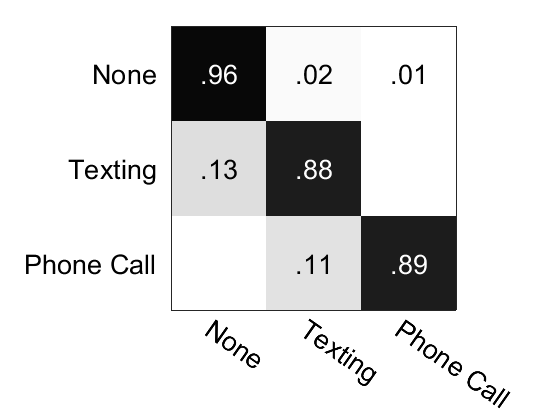}
	\end{subfigure}%
	~
	\begin{subfigure}[b]{0.22\textwidth}
		\centering
		\captionsetup{justification=centering}
		\hspace{0.1in} \caption{\ \ $K = 200$}
		\includegraphics[width=\textwidth]{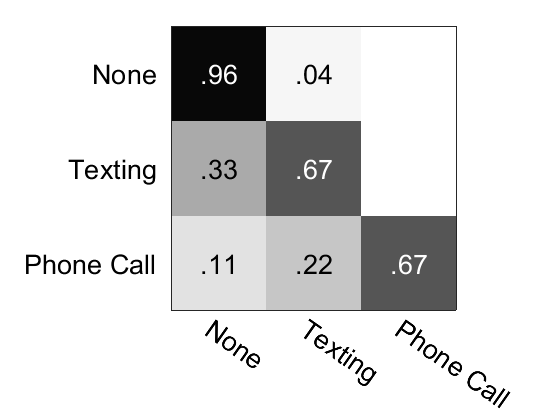}
	\end{subfigure}
	
\caption{Confusion matrices for MAP estimation (top row), and SVM based late fusion (bottom row). $K=100$ with SVM based late fusion results in the best overall accuracy.}
\label{fig:CM}
\end{figure*}

\section{Experimental Analysis}\label{experiments}

\subsection{Single Frame Activity Classification}

\begin{figure*}[t]
\begin{center}
\includegraphics[width=0.8\linewidth]{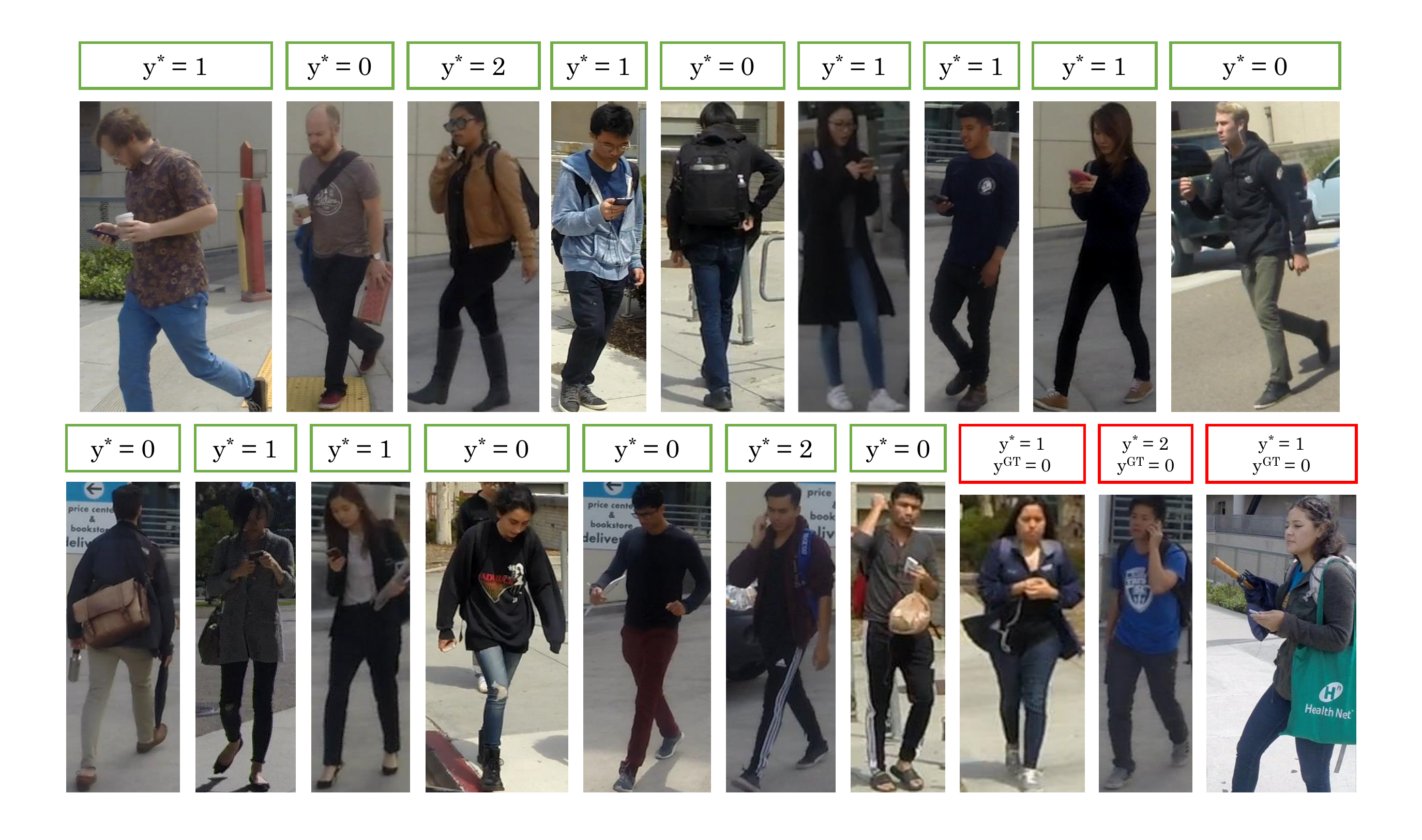}
\end{center}
\caption{Examples of pedestrians from the test set along with their predicted activity classes ($y^*$). Correctly predicted classes are enclosed in green boxes, and incorrectly predicted classes are enclosed in red boxes along with the corresponding ground truth class $y^{GT}$.}
\label{fig:results}
\end{figure*}

The critical hyperparamater that needs to be set for the proposed pipeline is the neighborhood size $K$. We experiment with a set of different values - $K \in \{25, 50, 100, 200\}$. Figure~\ref{fig:CM} shows confusion matrices for different values of $K$, for both the MAP estimation scheme and the SVM based late fusion. The MAP estimation scheme is seen to perform relatively poorly. This can be attributed to the fact that the pose prior $\prob(y_o; \mathcal{N}_K)$ is far too dominant in comparison to the other two probability terms. This leads to predictions that are overly influenced by the pose term, and hence the considerable false positives for all values of $K$. In comparison, the SVM weights each individual cue accordingly and predicts a more balanced output. This leads to  a much better overall accuracy in comparison to the MAP estimate. The best performance is seen for $K = 100$, which results in $94.6$\% overall accuracy. We notice that most mis-classifications in the output correspond to pedestrians who are considerably small in size ($\leq 200$ pixels in height), which makes it relatively harder to infer object labels and gaze information. Some other error modes are observed in cases where the pedestrian is holding objects that are considerably different to those observed in the training set. This issue however may be alleviated by collecting more data for training. Finally, there are cases where the correct label is ambiguous even to human annotators. This occurs when it is hard to infer the exact direction of gaze or the correct object label. For examplar results on the test dataset, we refer the reader to Figure~\ref{fig:results}.

\begin{table}[t!]
\centering
\tabcolsep=0.11cm
\caption{Per class and overall accuracies for four sets of cues - hands alone, pose alone, pose and hands, and pose, hands and gaze.}
 \begin{tabular}{|c c c c c|} 
 \hline
 \thead{Cues} & \thead{Accuracy \\for \textit{None}} & \thead{Accuracy \\for \textit{Texting}} & \thead{Accuracy \\for \textit{Phone Call}} & \thead{Overall \\Accuracy} \\ [0.5ex] 
 \hline\hline
 hand only & 0.94 & 0.58 & 0.20 & 0.810 \\ 
 pose only & 0.90 & 0.65 & 0.67 & 0.858 \\ 
 pose+hands & 0.93 & 0.71 & 0.81 & 0.916 \\
 pose+hands+gaze & \textbf{0.97} & \textbf{0.88} & \textbf{0.89} & \textbf{0.946} \\
 \hline
 \end{tabular}
 \label{table:pose_hand_head}
\end{table}

To understand the contribution and utility of each individual cue while making a prediction, we consider $K = 100$ corresponding to the best performing method. We train four separate SVM based fusion models, for four different sets of cues - hands only, pose only, pose and hands, and pose, hands and gaze. For each set of such cues, only features based on those cues are used for training the fusion SVM. Table~\ref{table:pose_hand_head} shows the per class and overall accuracies for each of these configurations. Pose alone is seen to perform reasonably well, indicating that it is the strongest of the three cues. However, it tends to be too harsh in its assignment, depending too heavily on the nearest neighbors. Adding hand related cues from the ensemble of ESVMs considerably improves the classification accuracy, especially for the \textit{texting} and \textit{phone call} classes as these are more reliant on recognition of hand-object interactions. Finally, adding gaze information further enhances the performance, noticeably for the \textit{texting} class since this requires the pedestrian to look at the phone directly.

As far as the processing time is concerned, our algorithm with $K = 100$ nearest neighbors takes about 4ms on average for each pedestrian on a 6th generation i7 CPU. This does not include the time for running the articulated pose estimation module, which we run independently on a Titan X GPU. As mentioned in \ref{speedup}, state-of-the-art pose estimation for multiple persons is almost nearing real time operation, and the processing times for other operations in our framework are minimal in comparison. This indicates feasibility for real world applications.

\subsection{Pose Tracking}

As tracking the pose enables us to achieve speedups during runtime, it is important to validate its reliability on pedestrian sequences. We do this by training the proposed GPDM based particle filter on three pedestrian sequences. The tracker is then evaluated on 4 separate sequences from the test set using the PCK metric~\cite{yang2013articulated}. The results are shown in Table~\ref{table:pose_tracking}. While tracking pedestrians in videos captured at 30Hz, it is seen that providing pose measurements even once every 6 frames (5Hz) is more than enough to result in very suitable tracks. Furthermore, tracking with a measurement at every frame provides an improvement over the tracks obtained by running the pose ConvNet alone. This illustrates that tracking gives us robust estimates of the pose in addition to making our algorithm run faster.

\begin{table}[t!]
\centering
\tabcolsep=0.09cm
\caption{PCK scores of proposed articulated pose tracking for different measurement rates, evaluated on 4 different pedestrian sequences. The tracking stays consistent and reliable even for relatively infrequent measurements obtained from the pose ConvNet.}
 \begin{tabular}{| c | c | c | c | c | c | c | c | c |} 
 \hline
 \thead{Measure-\\ment rate} & \thead{head} & \thead{neck} & \thead{right \\shoulder} & \thead{right \\elbow} & \thead{right \\wrist} & \thead{left \\shoulder} & \thead{left \\elbow} & \thead{left \\wrist} \\ [0.5ex] 
 \hline\hline
 untracked & 1.000 & 1.000 & 0.990 & 1.000 & 0.960 & 1.000 & 1.000 & 0.940\\ 
 30 Hz & 1.000 & 1.000 & 1.000 & 1.000 & 0.980 & 1.000 & 1.000 & 0.940\\ 
 15 Hz & 0.980 & 0.990 & 0.981 & 0.990 & 0.955 & 0.975 & 0.930 & 0.910\\ 
 10 Hz & 0.980 & 0.990 & 0.981 & 0.908 & 0.940 & 0.975 & 0.925 & 0.905\\ 
 5 Hz & 0.980 & 0.990 & 0.981 & 0.970 & 0.940 & 0.975 & 0.920 & 0.895\\ 
 \hline
 \end{tabular}
 \label{table:pose_tracking}
\end{table}

\subsection{Activity Classification for Sequential Data}

\begin{figure*}[t!]
    \centering
    \begin{subfigure}[b]{0.22\linewidth}
        \centering
        \includegraphics[height=1.2in]{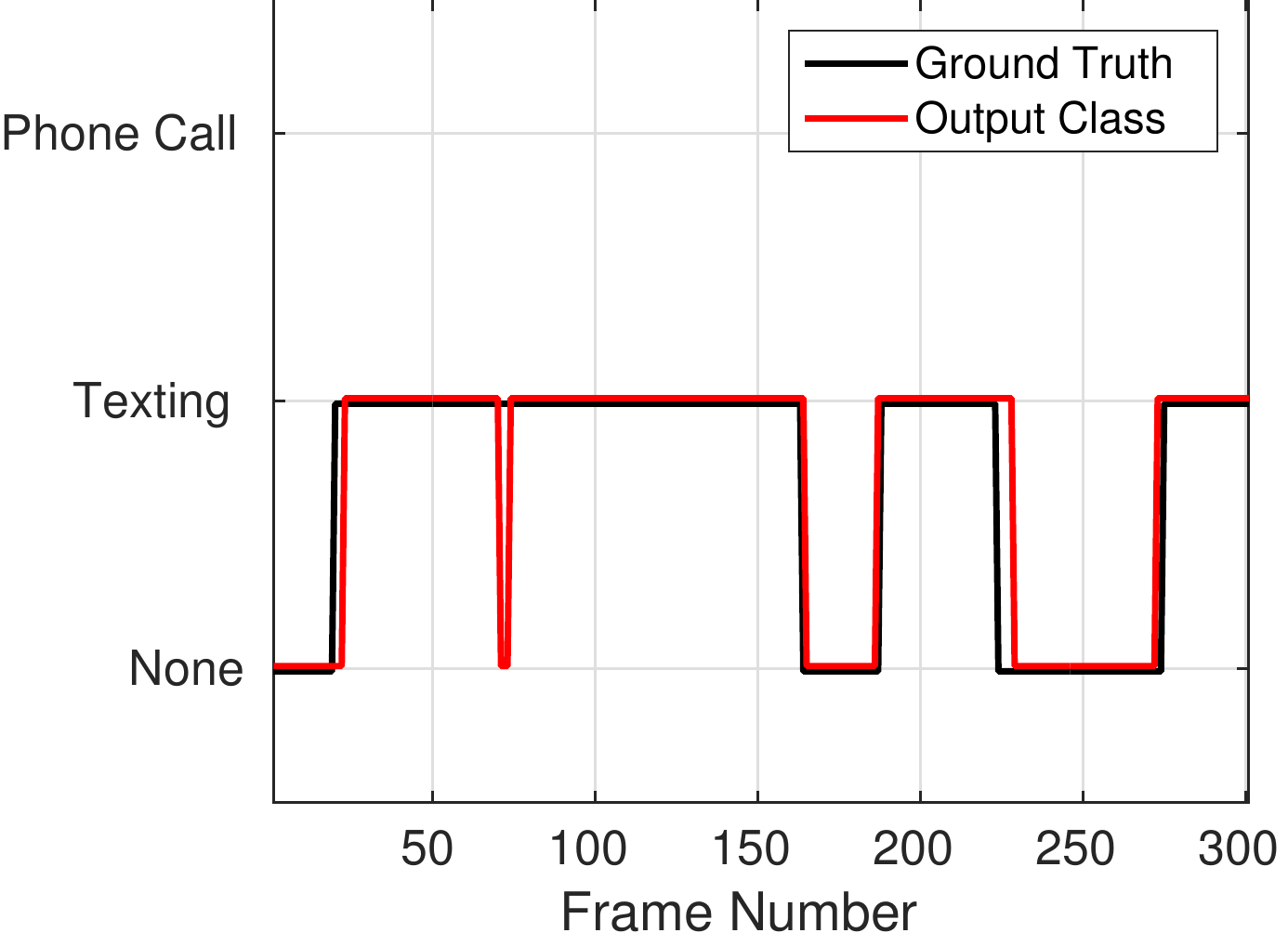}
        \caption{Sequence 1}
    \end{subfigure}
    ~
    \begin{subfigure}[b]{0.22\linewidth}
        \centering
        \includegraphics[height=1.2in]{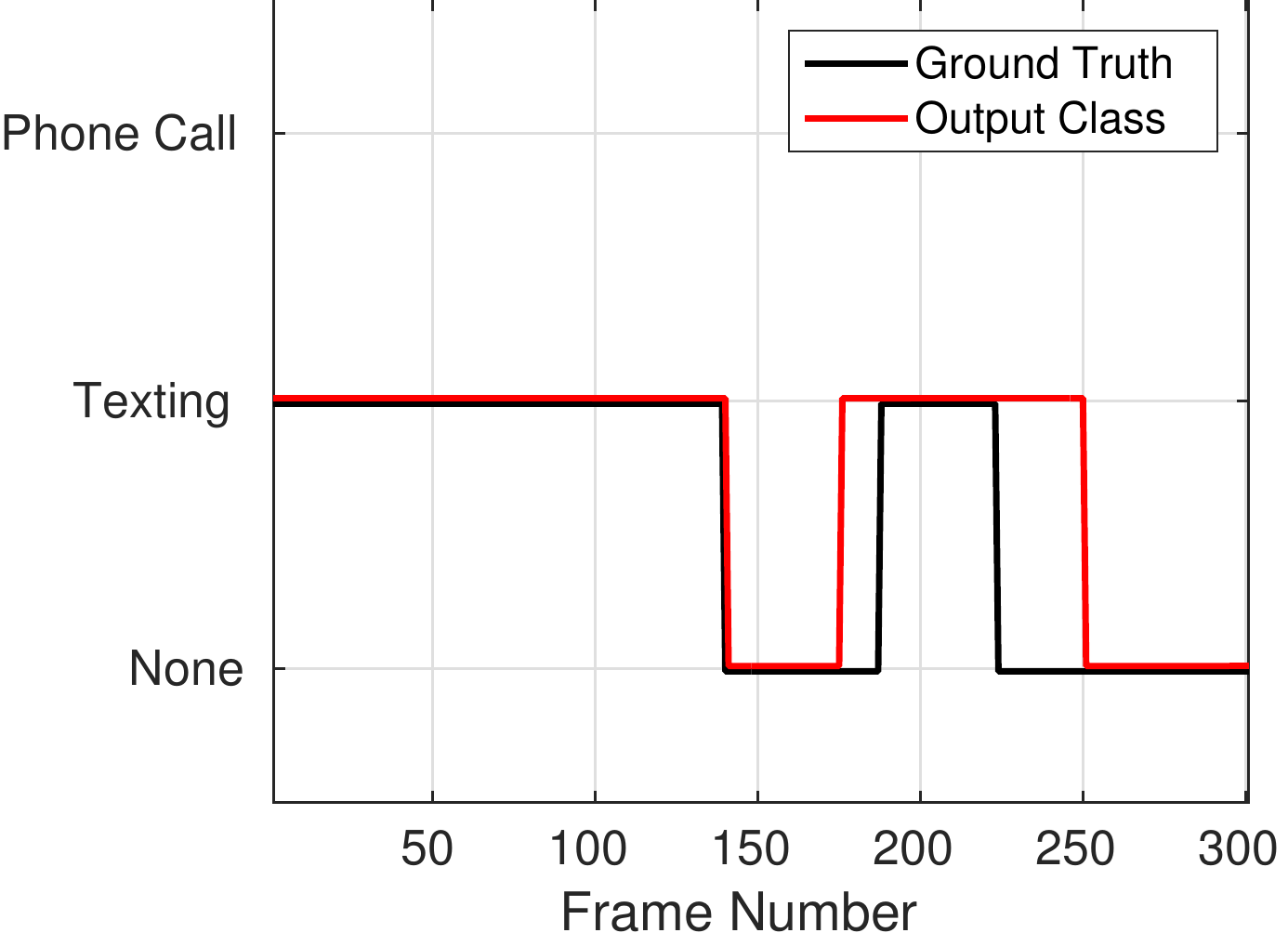}
        \caption{Sequence 2}
    \end{subfigure}
	~
    \begin{subfigure}[b]{0.22\linewidth}
        \centering
        \includegraphics[height=1.2in]{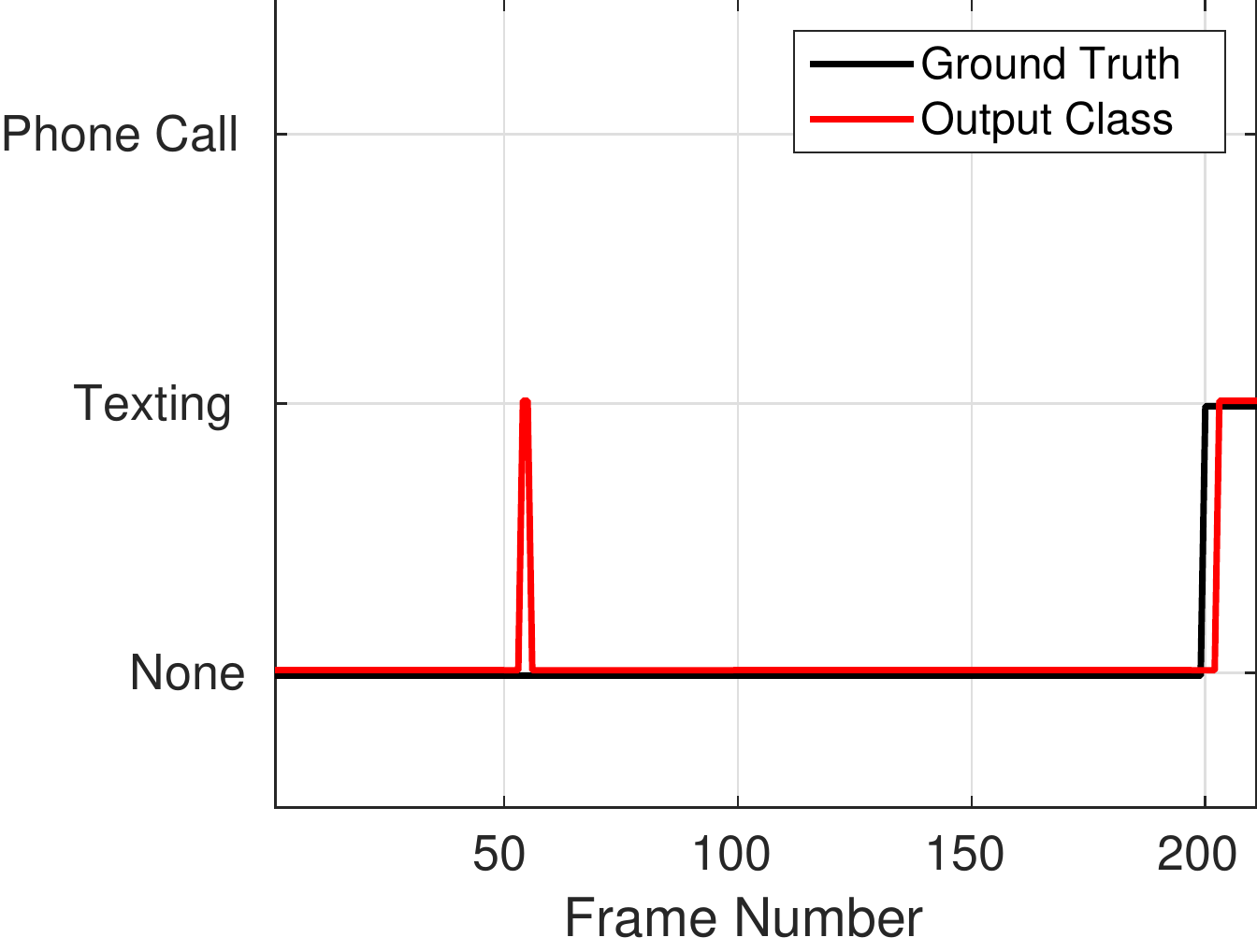}
        \caption{Sequence 3}
    \end{subfigure}
    ~
    \begin{subfigure}[b]{0.22\linewidth}
        \centering
        \includegraphics[height=1.2in]{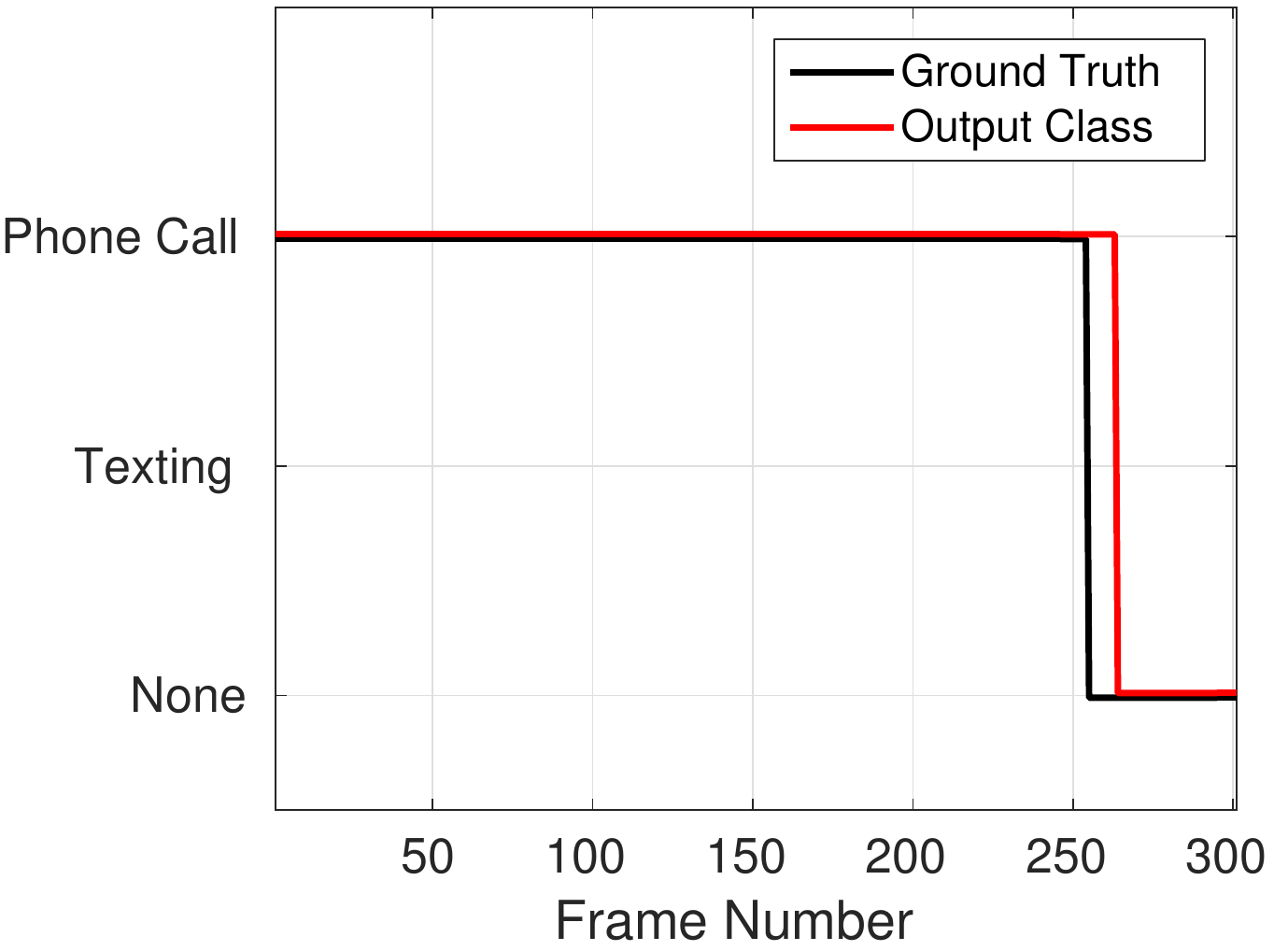}
        \caption{Sequence 4}
    \end{subfigure}
    \caption{Plot of ground truth and predicted output class as a function of frame number for 4 test sequences.}
\label{fig:sequential}
\end{figure*}

Next, we evaluate our proposed framework on 4 test sequences, each captured at 30 Hz and approximately 10 seconds in duration (280 - 310 frames per sequence). Each frame in a test sequence is annotated with the correct activity class. Our activity classification framework (with $K=100$) is run with the proposed articulated pose tracker, and predicts an output class for each frame. Additionally, we run the ensemble of ESVMs once every 50 frames as described in section \ref{speedup} to reduce the computational burden. We plot the predicted and ground truth classes as a function of the frame number for each of the 4 test sequences in Figure~\ref{fig:sequential}. It is clearly observable that the activity classification framework, along with the pose tracker result in class labels that are quite consistent with the ground truth, even under frequent changes in the activity dynamics.

\section{Concluding Remarks}\label{conclusions}

In this paper, we studied pedestrian distraction caused by cellphone usage in an effort to reduce growing number of pedestrian fatalities. To this end, a multi-cue pipeline to recognize pedestrian activity is proposed. A pedestrian is classified to be either \textit{texting}, in a \textit{phone call}, or be involved in no secondary activities based on cues from the articulated pose, hands and gaze. ESVMs trained offline are used to encode hand-object labels, whereas gaze features are obtained from a pre-trained ConvNet. Each cue is then used to propose scores based on the $K$ neighboring pedestrians from the training set. Finally, these scores are combined effectively using an SVM based late fusion scheme. In addition to this, we propose a GPDM based particle filter that operates based on measurements obtained from a pose estimation ConvNet in order to improve pose estimation, and speedup operation. Both the proposed methodology and the tracking framework are trained and evaluated on a unique \textit{pedestrian distraction} dataset, which provides rich semantic annotations to facilitate a more detailed study of pedestrians.

Although the results are promising, there are still many issues to be addressed. Pedestrian activities are rich in variety, and so are the objects they interact with. However, it must be noted that out proposal is highly scalable. Since it works on similarity based metrics obtained from pedestrians in the training set, as more diverse pedestrians are added to the training process, the performance would only improve, and without any evident drop in computational speed. Future work encompasses going beyond phone based distraction, and studying other sources of pedestrian distraction (e.g. talking, walking in a group, listening to music etc.), and integrating all such factors to predict a combined distraction score for each pedestrian. 

\section{Acknowledgments}
We would like to thank all our colleagues at the LISA lab, UCSD for their assistance in collecting and annotating the dataset. We would also like to express our gratitude to the reviewers and the editor for their valuable comments and suggestions.

\bibliographystyle{IEEEtran}
\bibliography{main_arxiv}

%
\begin{IEEEbiography}[{\includegraphics[width=1in,height=1.25in,clip,keepaspectratio]{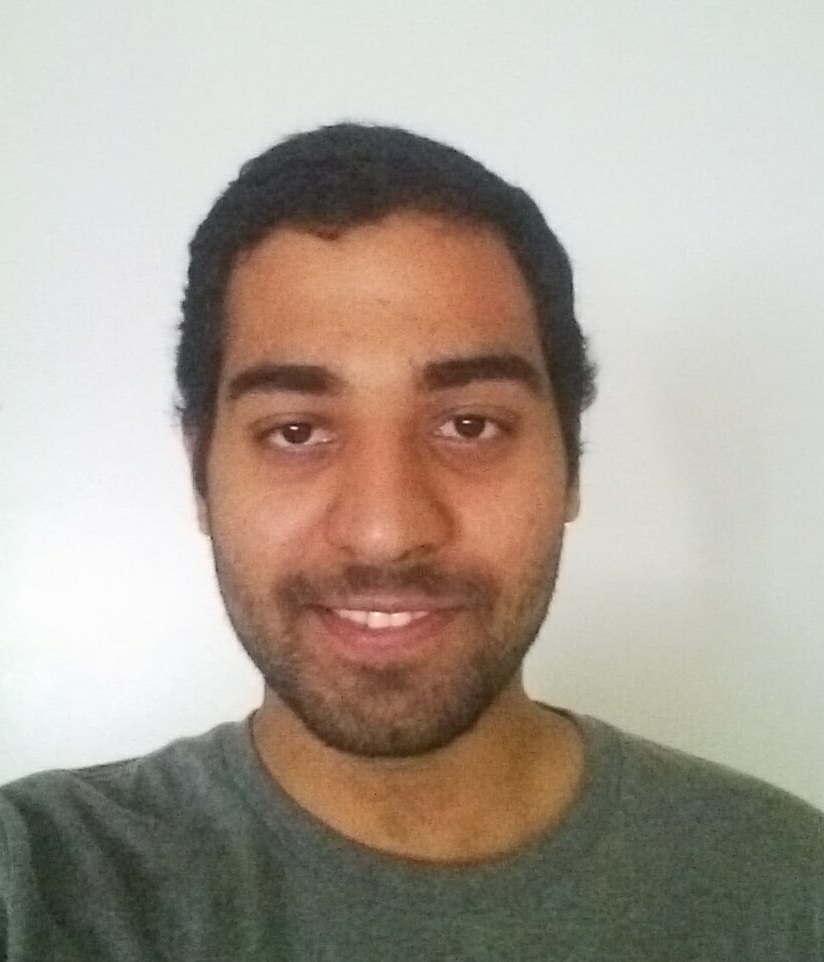}}]{Akshay Rangesh}
is currently working towards his PhD in electrical engineering from the University of California at San Diego (UCSD), with a focus on intelligent systems, robotics, and control. His research interests span computer vision and machine learning, with a focus on object detection and tracking, human activity recognition, and driver safety systems in general. He is also particularly interested in sensor fusion and multi-modal approaches for real time algorithms.
\end{IEEEbiography}

\begin{IEEEbiography}[{\includegraphics[width=1in,height=1.25in,clip,keepaspectratio]{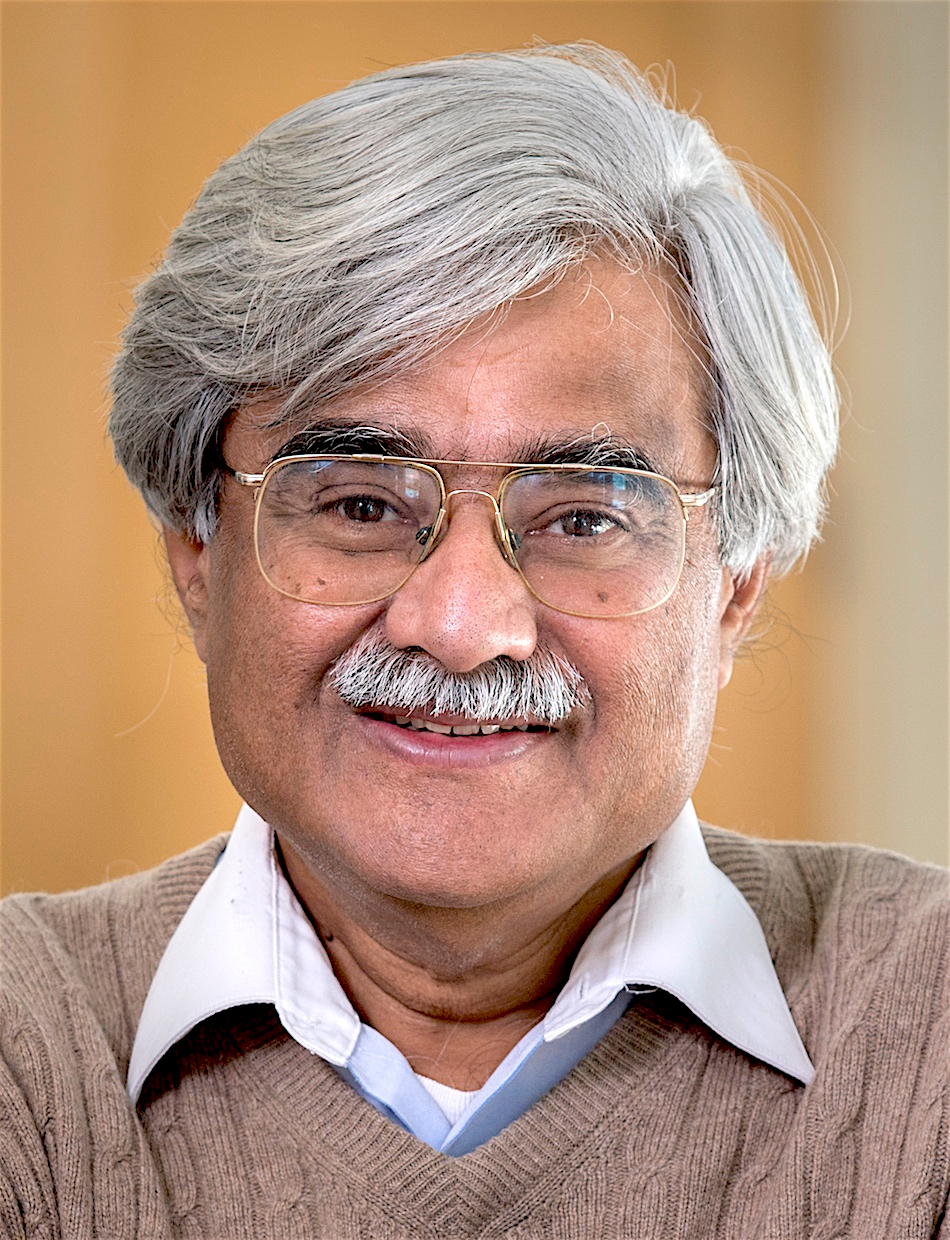}}]{Mohan Manubhai Trivedi}
is a Distinguished Professor at University of California, San Diego (UCSD) and the founding director of the UCSD LISA: Laboratory for Intelligent and Safe Automobiles,
winner of the IEEE ITSS Lead Institution Award (2015). Currently, Trivedi and his team
are pursuing research in intelligent vehicles, machine perception, machine learning, human-robot interactivity, driver assistance, active safety systems. Three of his students have received "best dissertation" recognitions. Trivedi is a Fellow of IEEE, ICPR and SPIE. He received the IEEE ITS Society's highest accolade "Outstanding Research Award" in 2013. Trivedi serves frequently as a consultant to industry and government agencies in the USA and abroad. 
\end{IEEEbiography}

\end{document}